# Linguistically Informed ChatGPT Prompts to Enhance Japanese-Chinese Machine Translation: A Case Study on Attributive Clauses

Wenshi Gu
*School of Foreign Languages,* Beihang Univeristy.
*Beijing, China*
*guwenshi@buaa.edu.cn*

## Abstract

*In the field of Japanese-Chinese translation linguistics, the issue of correctly translating attributive clauses has persistently proven to be challenging. Present-day machine translation tools often fail to accurately translate attributive clauses from Japanese to Chinese. In light of this, this paper investigates the linguistic problem underlying such difficulties, namely how does the semantic role of the modified noun affect the selection of translation patterns for attributive clauses, from a linguistic perspective. To address these difficulties, a pre-edit scheme is proposed, which aims to enhance the accuracy of translation. Furthermore, we propose a novel two-step prompt strategy, which combines this pre-edit scheme with ChatGPT, currently the most widely used large language model. This prompt strategy is capable of optimizing translation input in zero-shot scenarios and has been demonstrated to improve the average translation accuracy score by over 35%.*

## 1. Introduction

The development of machine translation has garnered significant attention in the industry. Essentially a complex task, machine translation (MT) [1] involves transforming a source input into semantically equivalent target output in a different language, requiring both sequence understanding and generation. Among the most recent and promising approaches to MT is neural machine translation (NMT) [2], which utilizes neural networks and handles massive datasets to learn from them. NMT is capable of generating more natural and fluent translations than its predecessors and is presently utilized in various commercial products such as Google Translation [3, 4] and Baidu Translation [5], among others. Nonetheless, NMT still faces several challenges, such as rare word handling, multi-language support, and quality assurance [6]. This paper tackles one specific domain of MT, namely Japanese-Chinese machine translation. This poses more challenges compared to languages with a large corpus such as English, especially in handling complex structures like attributive clauses, and is therefore more prone to errors [7].

Recently, OpenAI introduced ChatGPT [8], a large language model that is derived from the InstructGPT architecture [9]. The model is trained to follow specific prompts and deliver detailed responses. According to the official release, ChatGPT can address follow-up questions, identify and rectify erroneous responses, challenge false assumptions, and decline inappropriate requests, among other features, thanks to its conversational format. Its various natural language processing abilities include code debugging, logic reasoning, storytelling, machine translation, question answering, and more. Our interest lies in evaluating ChatGPT's performance in machine translation tasks. Furthermore, Large Language Models (LLMs) [10-12] like ChatGPT that are pretrained on massive, unlabeled corpora have displayed remarkable emergent abilities when subjected to model scaling, allowing for prompt-based downstream applications instead of task-specific fine-tuning. Prompting, which entails rephrasing test samples with descriptive task instructions and feeding the prompts directly to LLMs, is consequently being explored. A pertinent query arises: **Can we augment the prompt with *linguistic knowledge* to effectively enhance machine translation quality and improve model interpretability?**

In this paper, we try to answer the above question from a detailed but difficult point: Japanese to Chinese translation on attributive clause. In detail, when using a commercial product, i.e., machine translation site, the "internal related" attributive clause in (1-1) is translated as in (1-2).

（1-1）どちらかというと白羽さんが性犯罪者寸前の人間だと思っていたので、迷惑をかけられたアルバイト女性や女性客のことも考えずに、自分の苦しみの比喩として気軽に強姦という言葉を使う白羽さんを、被害者意識は強いのに、自分が加害者かもしれないとは考えない思考回路なんだなあ、と思って眺めた。

(『コンビニ人間』)

（1-2）？以为白羽先生快要成为性侵者了，所以没想到打工的女性和被打扰的女性顾客，随口说出我被强奸来比喻自己的苦难我看着用词的白羽先生，心想他虽然有很强的受害意识，却不认为自己可能是肇事者。

（Google Translation[1]）

In example (1), there is an extremely long attributive clause "迷惑をかけられたアルバイト女性や女性客のことも考えずに、自分の苦しみの比喩として気軽に強姦という言葉を使う（Casually using the word "rape" as a metaphor for personal suffering, without considering the female part-time workers or customers who may have been confused or negatively impacted by it.）"in front of the noun "白羽さん(Mr.Shiraha) ". The translation website struggles to accurately translate Japanese "attributive clause + modified noun" structures, such as example (1), into Chinese "attributive clause + modified noun" structures, leading to comprehension difficulties for Chinese native speakers. Specifically, the "Inner relation" Japanese attributive clause shown in (1-1) requires an accurate translation method that maintains comprehension for Chinese speakers. To address this issue, we propose a novel pre-edit scheme that considers the semantic role of the modified noun and explores the effectiveness of translating this type of Japanese attributive clause into Chinese from a linguistic perspective. Additionally, we incorporate prior linguistic knowledge into a common prompt as input for machine translation tools, resulting in improved translation performance. This approach paves a new path for combining linguistics and machine translation models.

[1] The machine translation site utilized in this study generated all translated texts, which were obtained on March 26, 2023.

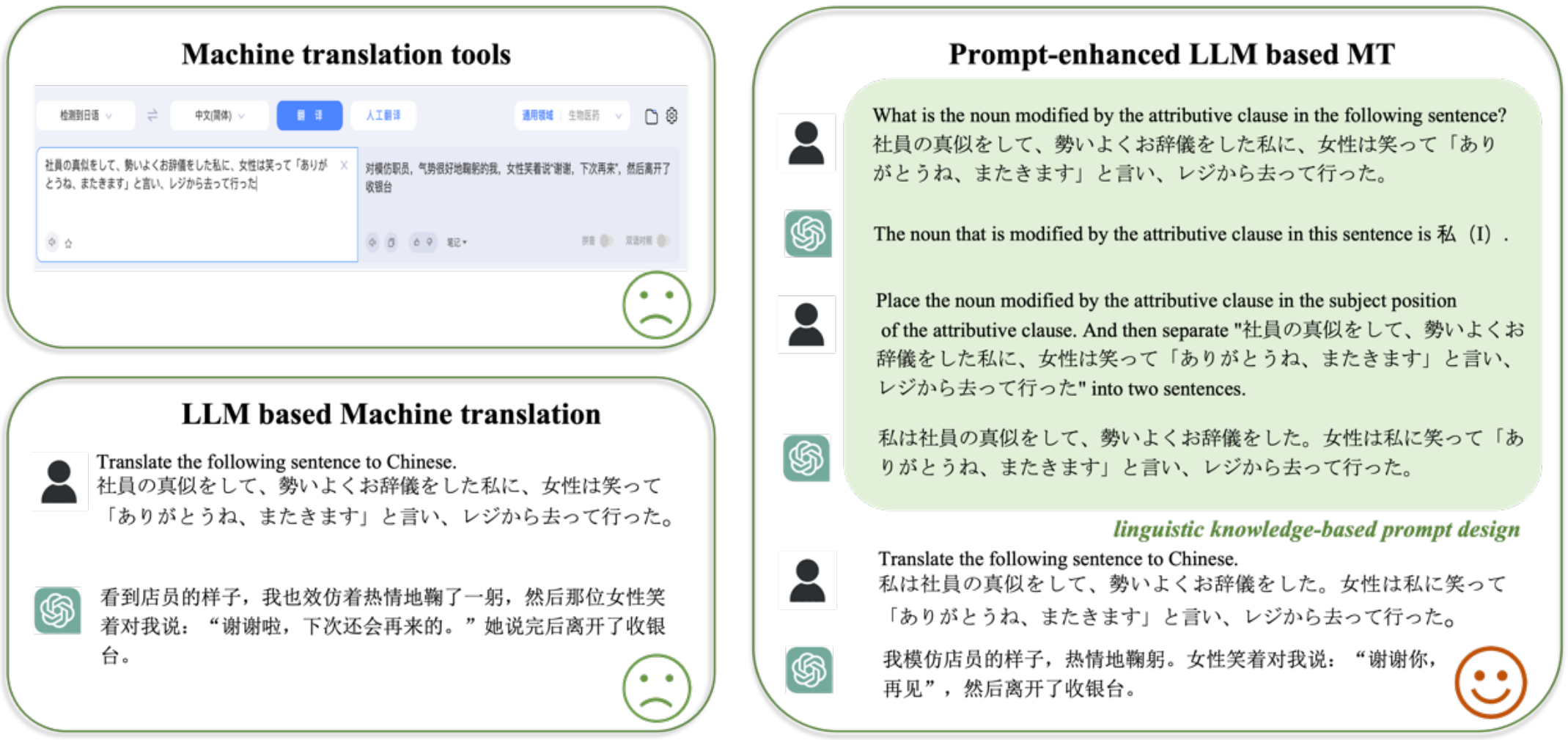

**Figure 1 Illustration of our proposed prompt-enhanced MT framework in comparison with conventional MT tools (e.g. Google translation) and vanilla LLM based MT.**

## 2. Related work

### 2.1 Previous research and our scope on linguistics

孔(2004) [13]proposed two translation methods for Japanese attributive clauses.

(2-1) The common practice for translating shorter attributive clauses is to directly render them as a attributive structure in Chinese.

(2-2) To translate longer attributive clauses, it is often necessary to relocate the attributive clause to the end of the main sentence.

孔(2004) proposed four specific translation methods for longer Japanese attributive clauses as mentioned in (2-2).

(2-3) Directly translating the attributive clause into Chinese without changing the modification relationship.

(2-4) Translate the attributive clause into a separate sentence and then translate the main clause.

(2-5) Translate the main clause first and then translate the attributive clause.

(2-6) Divide the attributive clause into two parts, use one part as a modifier before the modified noun, and translate the other part as a complement.

遠藤 (1990) [14] proposed three translation methods for longer Japanese attributive clauses.

(2-7) Long Japanese attributive clauses can also be translated into long attributive clauses in Chinese.

(2-8) Use long attributive clauses as predicates, with their centers serving as subjects, and then continue the sentence with appropriate words.

(2-9) Put the center word first (or along with a short attributive clause) and process the long attributive clause as a supplementary explanation that complements the center word with appropriate words (which may not always be necessary).

Although 孔(2004) and 遠藤(1990) proposed translation methods for Japanese attributive clauses, they did not elaborate on the characteristics of the "attributive clause + modified noun" structure in Japanese that each method is suitable for, nor did they discuss whether the semantic role of the modified noun influences the selection of a translation method.

The scoped object of this paper is an attributive clause that satisfies the following three conditions.

(2-10) "Inner relation" Japanese attributive clause

(2-11) The predicate in the main clause is a verb or a verb phrase.

(2-12) The nominative case is the only valid semantic role for the modified noun in the attributive clause under consideration.

According to (2-10), this paper does not focus on the "Outer relation" Japanese attributive clause, such as the one in (2-13). Additionally, as stated in (2-11) and (2-12), examples like (2-14) and (2-15), even if they are "Inner relation" Japanese attributive clauses, are excluded from the scope of this research. Therefore, only cases like (2-16), where the predicate in the main clause is a verb or verb phrase and the modified noun assumes a nominative semantic role in the attributive clause, are considered as the research objects.

(2-13) 太郎がテレビを壊した可能性がある。

(2-14) 太郎が壊したテレビは高い。

(2-15) 太郎が買ったテレビは壊れた。

(2-16) テレビを壊した太郎は頭を下げた。

Translated to English:

(2-13) There is a possibility that Taro broke the TV.

(2-14) The TV that Taro broke is expensive.

(2-15) The TV that Taro bought is broken.

(2-16) Taro who broke the TV bowed his head.

This paper refers to Japanese attributive clauses that satisfy the three conditions (2-10), (2-11), and (2-12) simultaneously, such as the example sentence (2-16), as "nominative" inner relation attributive clauses.

## 2.2 Large language model and prompting for MT

Figure 1 illustrates that conventional machine translation tools directly present the translated version of the original input sentence. In contrast, large language models (LLMs) require a specific template to be applied for formatting the original sentence to generate the translation, for instance, 'Translate the following sentence to Chinese: {original input sentence}'. In other words, given a pre-trained and fixed LLM, MT prompting first converts each test input to a prompt according to a template and then generates the translation by feeding the prompt to the LLM. This study primarily focuses on zero-shot prompting for translation, constraining itself to fast and direct implementation without any further fine-tuning or additional labeled examples. This approach also facilitates a fair comparison with conventional machine learning tools.

In the following section, we will initially present a comprehensive linguistic theory concerning the translation of Japanese "nominative" inner relation attributive clauses into Chinese. Subsequently, we will convert our devised theory into a standard, comprehendible prompt to enhance the quality of machine translation.

# 3. Linguistic theory

## 3.1 Translation patterns of Japanese "nominative" inner relation attributive clauses

When translating Japanese "nominative" inner relation attributive clauses into Chinese, the decision to adopt the "splitting long attributive clause and main clause into two separate sentences" translation method poses a challenge in determining whether the modified noun should initially fill in the valence complement of the verb in the attributive clause or the valence complement of the verb in the main clause. Prioritizing the valence complement of the verb in the attributive clause will result in the content of the attributive clause appearing before the main clause, while prioritizing the valence complement of the verb in the main clause will make the content of the main clause appear before that of the attributive clause.

If the modified noun gives priority to filling in the valence complement of the verb in the attributive clause, then the translation pattern is:

(3-1) Pattern I: (Inner relation attributive clause + modified noun) & Main clause (Japanese) => Attributive clause & modified noun, main clause (Chinese)

If the modified noun gives priority to filling in the valence complement of the verb in the main clause, then the translation pattern is:

(3-2) Pattern II: (Inner relation attributive clause + modified noun) & Main clause (Japanese) => Modified noun & main clause, Inner relation attributive clause (Chinese)

In (3-4), the translated sentence of (3-3), the modified noun "平介 (Heisuke)" gives priority to filling in the valence complement of "帰宅した (returned home)", which is the verb in the attributive clause. The content of this attributive clause "この日夜勤明けで、午前八時ちょうどに帰宅した (getting off work after the night shift, Heisuke returned home exactly at 8 o' clock in the morning)" comes before the main clause "四畳半の和室になり、テレビのスイッチを入れた (walked into a 4.5 tatami mat-sized Japanese-style room and turned on the television)". On the other hand, in (3-5), "平介 (Heisuke)", the modified noun, gives priority to filling in the valence complements of the verbs "なり (walked into)" and "入れた (turned on)", which are verbs in the main clause. The content of the main clause comes before that of the attributive clause.

(3-3)この日夜勤明けで、午前八時ちょうどに帰宅した平介は、四畳半の和室になり、テレビのスイッチを入れた。

（『秘密』）

(3-4)平介这一天下夜班，回到家刚好是上午八点，一走进四叠大小的和室，便打开了电视。

（Pattern I）

(3-5)这一天值完夜班，回到家刚好是早上八点，平介一走进四叠大小的和室，便打开了电视。

（Pattern II）

### 3.2 Exploring the relationship between the semantic role of modified nouns and translation pattern selection

How should we decide whether the modified noun should be placed in the valence complement of the attributive clause verb or the valence complement of the main clause verb? We propose a hypothesis: the priority order depends on the strength of the subordinate relationship between the semantic role of the modified noun and the verb. If the modified noun's semantic role has a stronger subordinate relationship with the attributive clause verb, it should be placed first in the valence complement of the attributive clause verb. Conversely, if the modified noun's semantic role has a stronger subordinate relationship with the main clause verb, it should be placed first in the valence complement of the main clause verb. If there is no difference in the strength of the subordinate relationship between the modified noun and the main clause verb and that with the attributive clause verb, the modified noun can be placed in either valence complement first.

仁田 (1980) [15]proposed that Japanese language represents various subordinate relationships between nouns and verbs with different semantic roles. Table 1, drawn from 仁田's research, lists the strong and weak subordinate relationships between each semantic role and verb. In this study, we will employ this table to examine and evaluate the strength of the relationship according to the classifications in the table.

**Table 1 Degree of Subordination of Verbs**

| | | | | | |
|---|---|---|---|---|---|
| strong subordinate relationships | case | high-level case elements | Agent | strong ↑ | Degree of Subordination of Verbs |
| | | | Patient | | |
| | | | Location (Existential Verbs) | | |
| | | low-level case elements | Source (Motion Verbs)<br>Goal (Motion Verbs)<br>Location (Space; Action Verbs)<br>Cause (Reason)<br>… | | |
| | | | Means | | |
| weak subordinate relationships | circumstantial element<br>adverbial element | | | ↓ weak | |

Note: Table 1 was compiled by the author based on 仁田's (1980: 23-32) research.

To validate the aforementioned hypothesis, this article extracted 64 instances of data pertaining to "nominative" inner relation attributive clauses from five Japanese novels, and examined the translation patterns of their respective Chinese versions. The analysis results have been summarized in Table 2.

**Table 2: Semantic Role Relationship between Modified Noun and Attributive/Main Clause Verb in Inner Relation Attributive Clauses**

| | | Semantic Role Relationship between Modified Noun and Main Clause Verb | Pattern II | Pattern I | Others | Total |
|---|---|---|---|---|---|---|
| Semantic Role Relationship between Modified Noun and Attributive Clause Verb | Agent | Nominative case | 3 | 21 | 0 | 24 |
| | | Accusative case | 0 | 7 | 14 | 21 |
| | | Dative case | 2 | 9 | 0 | 11 |
| | | Causative case | 0 | 4 | 0 | 4 |
| | | Allative case | 0 | 1 | 0 | 1 |
| | | Locative case | 0 | 1 | 0 | 1 |
| | | Adverbial | 0 | 2 | 0 | 2 |
| Total | | | 5 | 45 | 14 | 64 |

When the semantic role relationship between the modified noun and the verb in the main clause is "nominative," the strength of the subordinate relationship between the semantic role of the modified noun and the verb in the main clause and the subordinate relationship between the semantic role of the modified noun and the verb in the attributive clause are equal. It is abbreviated as "attributive clause verb = main clause verb." When the semantic role relationship between the modified noun and the verb in the main clause is other than nominative, then the subordinate relationship between the semantic role of the modified noun and the verb in the main clause is weaker than that between the semantic role of the modified noun and the verb in the attributive clause. It is abbreviated as "attributive clause verb > main clause verb." The correlation between the degree of subordination between semantic roles attributed to modified nouns and verbs and the selection of translation mode is presented in Table 3.

**Table 3 Degree of Subordination for Attributive/Main Clause Verbs**

| | attributive clause verb > main clause verb | attributive clause verb = main clause verb |
|---|---|---|
| Pattern I | 24（60%） | 21（87.5%） |
| Pattern II | 2（5%） | 3（12.5%） |
| Others | 14（35%） | 0（%） |
| Total | 40（100.00%） | 24（100.00%） |

Based on the results presented in Table 3, it is evident that Pattern I has a significantly higher number of translations than Pattern II. This supports the previously proposed hypothesis that the choice of translation mode for "nominative" inner relation attributive clauses is linked to the degree of subordination between the semantic role of the modified noun and the verb. As "nominative" inner relation attributive clauses have a "nominative" semantic relationship, which represents the highest level of subordinate attention intensity, when translating into Chinese, it is recommended that priority be given to using the modified noun as the valence complement of the attributive clause verb - nominative.

# 4. Prompt-enhanced Machine Translation

Section 3 of this paper yielded a crucial finding (4-1) regarding the translation of Japanese "nominative" inner relation attributive clauses into Chinese. The study concluded that priority should be given to using the modified noun to fill the nominative case of the attributive clause verb and placing the content of the attributive clause before the content of the main clause to ensure efficient translation. In this chapter, we employ large language models to investigate and validate the feasibility of this conclusion.

## 4.1 Machine Translation Linguistic Metric

To accurately assess the quality of machine translation, it is essential to establish comprehensive evaluation criteria. Building on 宮田&藤田's (2017) [16]Japanese-English machine translation evaluation framework, we have developed Japanese-Chinese machine translation evaluation criteria that take into account the unique characteristics of Japanese-Chinese translation. The resulting evaluation framework is presented in Table 4 for a more intuitive and informative reflection of machine translation quality.

**Table 4 Japanese-Chinese machine translation translation quality evaluation criteria**

| | |
|---|---|
| 5. | The original information is complete and accurate, with no omissions or mistranslations, and no grammatical errors. The translated content is natural and fluent, and is identical to the original content. Native Chinese speakers can easily understand the translation and obtain semantic information that is essentially the same as in the original text. |
| 4. | The original information is error-free, with no omissions or mistranslations, and no grammatical errors. The translated content closely resembles the original content in meaning and is also clear and natural. Native Chinese speakers can easily understand the translation and obtain semantic information that is essentially the same as in the original text. |
| 3. | The original information is largely complete and accurate, with few omissions or mistranslations, and few or no grammatical errors. The translated content is similar to the original content in meaning and is mostly clear and easy to understand. Native Chinese speakers can comprehend the translation with relative ease and obtain most of the semantic information in the original text. |
| 2. | The original information contains serious omissions or |
| | mistranslations, or major grammatical errors. The translated content significantly differs from the original content in meaning and is difficult to understand. Native Chinese speakers can comprehend the translation with difficulty but cannot obtain the same semantic information as in the original text. |
| 1. | The translation is incomprehensible to native Chinese speakers and fails to convey the intended meaning of the original text. |

## 4.2 Experimental results

Although the linguistic conclusion is valuable in enhancing translation quality, it may not be immediately applicable to improve the output of translation websites like Google and Baidu. It's because these websites operate solely in the "input-output" mode, with no provision for users to debug the model. Fortunately, large language models like ChatGPT are interactive translation tools that function within a dialogue-based framework, allowing users to iteratively obtain superior outputs by adding prompts. This paper aims to convert the linguistic conclusion derived in Section 3 into a prompt and evaluate its effectiveness by comparing the translation quality before and after implementing the prompt.

The recommendation for translating "nominative" inner relation attributive clauses into Chinese is to prioritize using the modified nouns to fill the valence complement of the attributive clause verbs, which means filling the nominative case of the attributive clause verbs with the modified nouns. Additionally, "the content of the attributive clause is placed before the content of the main clause" can be transformed into prompt:

(4-2) ***To rephrase the following sentence, place the nouns modified by the attributive clause in the subject position of the original clause and restructure the sentence.***

To assess the efficacy of the prompt, we selected five corpora featuring Japanese "nominative" inner relation attributive clauses and conducted translation experiments using ChatGPT. The experimental procedure entailed the following steps:

(4-3a) Question:

*Translate the following sentence into Chinese*: 社員の真似をして、勢いよくお辞儀をした私に、女性は笑って「ありがとうね、またきます」と言い、レジから去って行った。

(4-3b) Answer:

看到店员的样子，我也效仿着热情地鞠了一躬，然后那位女性笑着对我说：“谢谢啦，下次还会再来的。”她说完后离开了收银台。

(4-4a) Question:

*What is the noun modified by the attributive clause in the following sentence?*社員の真似をして、勢いよくお辞儀をした私に、女性は笑って「ありがとうね、またきます」と言い、レジから去って行った。

(4-4b) Answer:

*The noun that is modified by the attributive clause in this sentence is* 私（I）.

(4-5a) Question:

*Please place the noun modified by the attributive clause in the subject position of the attributive clause, and then separate* "社員の真似をして、勢いよくお辞儀をした私に、女性は笑って「ありがとうね、またきます」と言い、レジから去って行った" *into two sentences.*

(4-5b) Answer:

私は社員の真似をして、勢いよくお辞儀をした。女性は私に笑って「ありがとうね、またきます」と言い、レジから去って行った。

(4-6a) Question:

*Translate the following sentence into Chinese.* 私は社員の真似をして、勢いよくお辞儀をした。女性は私に笑って「ありがとうね、またきます」と言い、レジから去って行った。

(4-6b) Answer:

我模仿店员的样子，热情地鞠躬。女性笑着对我说：“谢谢你，再见”，然后离开了收银台。

By adding a prompt, the Japanese translation of “社員の真似をして、勢いよくお辞儀をした私に、女性は笑って「ありがとうね、またきます」と言い、レジから去って行った” changed from “看到店员的样子，我也效仿着热情地鞠了一躬，然后那位女性笑着对我说：“谢谢啦，下次还会再来的。”她说完后离开了收银台” (4-3b) to “我模仿店员的样子，热情地鞠躬。女性笑着对我说：“谢谢你，再见”，然后离开了收银台” (4-6b). The translation quality has also been improved from 3 to 4.

This paper conducted the above experiments on 5 pieces of data, and the improvements in translation quality are shown in Table 5.

**Table 5: Translation Quality Changes Before and After Adding Prompt**

| Source Language (SL) | ChatGPT | |
|---|---|---|
| | Target Language1 (TL1) (Before adding prompt) | Target Language 2 (TL2) (After adding prompt) |
| 1 | 3 | 5 |
| 2 | 4 | 4 |
| 3 | 3 | 4 |
| 4 | 3 | 5 |
| 5 | 3 | 4 |
| Average | 3.2 | 4.4 (37.5%) |

Table 5 demonstrates that by integrating linguistic prompts into ChatGPT, the average translation quality improved significantly from 3.2 to 4.4, marking a notable 37.5% increase. Moreover, Examples (4-7a), (4-7b), and (4-7c) illustrate that TL1 incorporated linguistic details absent in SL, such as "I saw the behavior of the clerk," whereas TL2 reflected language systems akin to that of SL, showing no signs of over- or under-translation, or mistranslation. Additionally, TL2 features a subject placement consistent with Chinese expression norms, rendering the language more natural, fluent, and concise compared to TL1.

(4-7a) 社員の真似をして、勢いよくお辞儀をした私に、女性は笑って「ありがとうね、またきます」と言い、レジから去って行った。

(SL)

(4-7b) 看到店员的样子，我也效仿着热情地鞠了一躬，然后那位女性笑着对我说：“谢谢啦，下次还会再来的。”她说完后离开了收银台。

(TL1)

(4-7c) 我模仿店员的样子，热情地鞠躬。女性笑着对我说：“谢谢你，再见”，然后离开了收银台。

(TL2)

## 5. Conclusion

In this paper, we propose a new approach for enhancing the accuracy of machine translation models through the combination of linguistic knowledge and neural methods. By establishing a clear link between the semantic roles of modified nouns and the selection of translation modes in Japanese attributive clauses, we present new insights into the translation process and underscore the importance of incorporating linguistic cues to improve translation accuracy. Leveraging this linguistic knowledge, we generate prompts that are specifically designed to aid machine translation and conduct experiments utilizing ChatGPT to verify the practicality of our approach. The results obtained demonstrate the effectiveness of our approach in improving the quality of translations, thus highlighting the potential value of incorporating linguistic knowledge into machine translation systems. Our experiments reveal a significant improvement in machine translation accuracy, with more than 30% improvement on the linguistic side. These results emphasize the promising direction of incorporating linguistic knowledge and large language models to improve the quality of machine translation applications.

## Acknowledgments

This work is supported by Beijing Social Science Foundation （No.KZ31082901） and Beihang Univeristy Sponsored Projects for Core Young Researchers in the Disciplines of Social Sciences and Humanities （No.KG16183601）.

## Appendix 1

| ChatGPT | | | | | |
|---|---|---|---|---|---|
| Source Language1（Before adding prompt） | Target Language1（Before adding prompt） | Translation Quality 1 | Source Language2（After adding prompt） | Target Language 2（After adding prompt） | Translation Quality 2 |
| 1 四日夜に啓子を殺害した梶は、その三日後の七日朝、つまり昨日になって中央署に自首してきた。[2] | 杀害了啓子的梶在四日晚上，三天后的七日早上即昨天到中央署自首了。 | 3 | 梶は四日夜に啓子を殺害した。梶はその三日後の七日朝、つまり昨日になって中央署に自首してきた。 | 梶在四日晚上杀害了啓子。他在三天之后的七日早上即昨天自首到中央警署。 | 5 |
| 2 この日夜勤明けで、午前八時ちょうどに帰宅した平介は、四畳半の和室に入るなり、テレビのスイッチを入れた。[1] | 这天夜班结束后，平介早上八点准时回到家，一进入四畳半的和室就打开了电视开关。 | 4 | 平介は、この日夜勤明けで、午前八時ちょうどに帰宅した。彼は四畳半の和室に入るなり、テレビのスイッチを入れた。 | 平介在这个日夜班结束后早上八点准时回到家，一进入四畳半的和室就打开了电视开关。 | 4 |
| 3 ユカリが駅前のショッピングモールで買ったケーキを食べながら、皆の顔を見て懐かしい懐かしいと連呼すると、皆は笑った。[3] | 一边吃着在车站前购买的蛋糕，一边看着大家的脸，不停地喊着“好怀念，好怀念”的 Yukari 让大家都笑了。 | 3 | ユカリが駅前のショッピングモールで買ったケーキを食べながら、皆の顔を見て懐かしい懐かしいと連呼するのに皆は笑った。 | 当 Yukari 一边吃着她在车站前的购物中心买的蛋糕，一边看着每个人的脸并不停地说着“好怀念，好怀念”的时候，大家都笑了。 | 4 |
| 4 身につけている洋服も、発する言葉のリズムも変わってしまった私が笑っている。[3] | 身上穿着的衣服和说话的语调都变了的我，却在笑着。 | 3 | 私が身につけている洋服も、発する言葉のリズムも変わってしまったが、笑っている。 | 我身上的衣服和说话的节奏都变了，但我还是在笑。 | 5 |
| 5 社員の真似をして、勢いよくお辞儀をした私に、女性は笑って「ありがとうね、またきます」と言い、レジから去って行った。[3] | 看到店员的样子，我也效仿着热情地鞠了一躬，然后那位女性笑着对我说：“谢谢啦，下次还会再来的。”她说完后离开了收银台。 | 3 | 私は社員の真似をして、勢いよくお辞儀をした。女性は私に笑って「ありがとうね、またきます」と言い、レジから去って行った。 | 我模仿店员的样子，热情地鞠躬。女性笑着对我说：“谢谢你，再见”，然后离开了收银台。 | 4 |

## Appendix 2 Sources of Example Sentences

[1]東野圭吾『秘密』東京：文藝春秋(1998)

[2]横山秀夫『半落ち』東京：講談社(2002)

[3]村田沙耶香『コンビニ人間』東京：文藝春秋(2016)